\documentclass[letterpaper, 10 pt, conference]{ieeeconf}  
\usepackage{blindtext}
\usepackage{textcomp}
\usepackage{fancyhdr}
\usepackage{indentfirst}
\usepackage{colortbl} 
\usepackage[table]{xcolor} 
\usepackage{graphicx}
\usepackage{mathtools}
\usepackage{psfrag}
\usepackage{pstricks}
\usepackage{amssymb}
\usepackage{algorithm2e}
\usepackage{nomencl}
\usepackage{tikz}
\usetikzlibrary{shapes,arrows}
\usepackage{amsmath}
\usepackage{amssymb}

\usepackage{floatflt}
\usepackage{url}
\usepackage{afterpage}
\usepackage[bb=boondox,bbscaled=.95]{mathalfa}
\usepackage[english]{babel}

\usepackage{pgfplots}
\usepgfplotslibrary{groupplots}
\usepgfplotslibrary{patchplots}
\pgfplotsset{
	compat=1.14,
	colormap={no data}{
		color=(white)
		color=(red)
	},
	colormap/bluered,
	colormap={parula}{
		rgb255=(53,42,135)
		rgb255=(15,92,221)
		rgb255=(18,125,216)
		rgb255=(7,156,207)
		rgb255=(21,177,180)
		rgb255=(89,189,140)
		rgb255=(165,190,107)
		rgb255=(225,185,82)
		rgb255=(252,206,46)
		rgb255=(249,251,14)
	},
	colormap={blackwhite}{gray(0cm)=(0.5); gray(1cm)=(1)},
}

\usepackage{tikz}
\usetikzlibrary{shapes,arrows,fit}

\usetikzlibrary{external}
\tikzexternaldisable
\tikzset{external/system call={%
  pdflatex \tikzexternalcheckshellescape -halt-on-error -interaction=batchmode -jobname "\image" "\texsource"; 
  ps2pdf13 "\image".pdf "\image-13".pdf && cp "\image-13".pdf "\image".pdf}}

\DeclareMathOperator{\R}{\mathbb{R}}
\DeclareMathOperator*{\argmin}{arg\,min}

\DeclareMathOperator{\Ha}{\mathcal{H}}

\DeclareMathOperator{\X}{\mathcal{X}}
\DeclareMathOperator{\V}{\mathcal{V}}
\DeclareMathOperator{\D}{\mathcal{D}}
\DeclareMathOperator{\U}{\mathcal{U}}

\DeclareMathOperator{\Y}{\mathcal{Y}}
\DeclareMathOperator{\J}{\mathcal{J}}
\DeclareMathOperator{\C}{\mathcal{C}}

\DeclareMathOperator{\K}{\mathcal{K}}

\newtheorem{thm}{Theorem}[section]

\newtheorem{assum}[thm]{Assumption}

\newtheorem{defn}[thm]{Definition}
\newtheorem{exmp}[thm]{Example}
\newtheorem{rem}[thm]{Remark}

\IEEEoverridecommandlockouts                              

\title{\LARGE \bf
Port--Hamiltonian Approach to Neural Network Training
}

\author{Stefano Massaroli$^{1,\dagger,\star}$ and Michael Poli$^{2,\star}$,\\ Federico Califano$^{3}$, Angela Faragasso$^{1,\dagger}$, Jinkyoo Park$^{2}$,
Atsushi Yamashita$^{1,\dagger}$ and Hajime Asama$^{1,\ddagger}$
\thanks{$^{1}$Stefano Massaroli, Angela Faragasso, Atsushi Yamashita and Hajime Asama are with the Department of Precision Engineering, The University of Tokyo, 7-1-3 Hongo, Bunkyo, Tokyo, Japan \newline {\tt\small \{massaroli,faragasso,yamashita,asama\} @robot.t.u-tokyo.ac.jp}}%
\thanks{$^{2}$Michael Poli and Jinkyoo Park are with the Department of Industrial and Systems Engineering, Korea Advanced Institute of Science and Technology (KAIST), 291 Daehak-ro, Eoeun-dong, Yuseong-gu, Daejeon, South Korea
        {\tt\small \{poli\_m,jinkyoo.park\}@kaist.ac.kr}}%
\thanks{$^{3}$Federico Califano is with the Faculty of Electrical Engineering, Mathematics \& Computer Science (EWI), Robotics and Mechatronics (RAM), University of Twente, Hallenweg 23 7522NH, Enschede, The Netherlands {\tt f.califano@utwente.nl}}%
\thanks{$^{\dagger}$\textit{IEEE Member}, $^\ddagger$\textit{IEEE Fellow}}%
\thanks{$^{\star}$These authors contributed equally to the work.}%
\thanks{\textcopyright~ \textbf{2019 IEEE.  Personal use of this material is permitted.  Permission from IEEE must be obtained for all other uses, in any current or future media, including reprinting/republishing this material for advertising or promotional purposes, creating new collective works, for resale or redistribution to servers or lists, or reuse of any copyrighted component of this work in other works.}}
}

\begin{document}

\maketitle
\thispagestyle{empty}
\pagestyle{empty}

\setlength{\belowdisplayskip}{5pt} \setlength{\belowdisplayshortskip}{5pt}
\setlength{\abovedisplayskip}{5pt} \setlength{\abovedisplayshortskip}{5pt}
\begin{abstract}
Neural networks are discrete entities: subdivided into \textit{discrete} layers and parametrized by weights which are iteratively optimized via \textit{difference} equations. Recent work proposes networks with layer outputs which are no longer quantized but are solutions of an ordinary differential equation (ODE); however, these networks are still optimized via discrete methods (e.g. gradient descent). In this paper, we explore a different direction: namely, we propose a novel framework for learning in which the parameters themselves are solutions of ODEs. By viewing the optimization process as the evolution of a port-Hamiltonian system, we can ensure convergence to a minimum of the objective function. Numerical experiments have been performed to show the validity and effectiveness of the proposed methods.
\end{abstract}
%
%
\section{Introduction}
Neural networks are universal function approximators \cite{hornik1989multilayer}. Given enough \textit{capacity}, which can arbitrarily be increased by adding more parameters to the model, they can approximate any Borel--measurable function mapping finite--dimensional spaces. Each layer of a neural network performs an affine transformation to its input and generates an output which is then fed into the next layer. Backpropagation \cite{rumelhart1985learning} is at the core of modern deep learning, and most state-of-the-art architectures for tasks such as image segmentation \cite{he2017mask}, generative tasks \cite{brock2018large}, image classification \cite{he2016deep} and machine translation \cite{devlin2018bert} rely on the effective combination of universal approximators and line search optimization methods: most notably \textit{stochastic gradient descent} (SGD), Adam \cite{KingmaB14} RMSProp \cite{tieleman2012lecture} and recently RAdam \cite{liu2019variance}.

Training neural networks is a non--convex optimization problem which aims to obtain globally or locally optimal values for its parameters by minimizing an objective function that is usually designed ad--hoc for the application at hand. The landscape of such objective functions is often highly non--convex and finding global optima is in general an NP--complete problem \cite{li2018visualizing,blum1989training}. Optimality guarantees for algorithms such as gradient descent do not hold in this setting; moreover, the discrete nature of neural networks adds complications to the development of a proper theoretical understanding with sufficient convergence conditions. Despite the empirical successes of deep learning, these reasons alone lead many to question whether or not relying on these standard methods could be a limitation to the advancement of deep learning research. In this work, we offer a new perspective on the optimization of neural networks, where parameters are no longer iteratively updated via difference equations, but are instead solutions of ODEs. This is achieved by equipping the parameters with autonomous port-Hamiltonian dynamics. 
\begin{figure}
    \centering
    \definecolor{ocean}{rgb}{0.00000,0.44700,0.74100}
\begin{tikzpicture}
\begin{axis}[
width=5cm,
height=5cm,
at={(1in,0.331in)},
view={0}{90},
colormap/viridis,
xmin=-2,
xmax=2,
xlabel style={at = {(0.5,0)}},
xlabel={$\vartheta^{(1)}$},
ymin=-1,
ymax=1,
ylabel style={at = {(0,0.5)}},
ylabel={$\vartheta^{(2)}$},
yticklabels={,,},
xticklabels={,,},
title = {Gradient Descent}
]
\addplot3[contour filled={number = 50,labels={false}},mesh/rows=50,mesh/cols=50,mesh/check=false,forget plot
] table {H.dat};
\addplot [color=white, line width=1pt, dotted]
table[row sep=crcr]{%
1.9	1\\
0.684043000000002	-0.485\\
0.294304273625746	-0.89580455\\
0.106966322024757	-0.289669633009248\\
0.0235950766583691	-0.594984356941613\\
0.0845451874151783	-0.806997019564963\\
0.104900662253661	-0.526750133142011\\
0.0594830483179512	-0.823813162153181\\
0.111940327014499	-0.479481649875851\\
0.0512962904368019	-0.807088831185495\\
0.110974030319992	-0.521534234332979\\
0.0577522856694644	-0.823566428367056\\
0.112227019021586	-0.479884502657897\\
0.0513129262891225	-0.807350708833395\\
0.111010150143497	-0.520884249218682\\
0.0576492376976389	-0.823413214603245\\
0.112223351400711	-0.480280052953319\\
0.0513728186251655	-0.807563973858598\\
0.111030757598379	-0.520361284196152\\
0.0575676268698583	-0.823286373091954\\
0.112219624895777	-0.480607862660636\\
0.051422558872083	-0.807739792856561\\
0.111047678569761	-0.519929800571638\\
0.0575003048971759	-0.823180164456706\\
0.112216316742346	-0.480882353983539\\
0.0514642375165078	-0.807886414586435\\
0.111061752594008	-0.519569725822114\\
0.0574441317900204	-0.823090458931686\\
0.112213395102049	-0.481114192832599\\
0.0514994593013675	-0.808009827604124\\
0.111073572685886	-0.519266474422042\\
};
\addplot [draw=none, mark=*, mark options={solid, fill=orange}, mark size=2pt]
table[row sep=crcr]{%
1.9	1\\
0.684043000000002	-0.485\\
0.294304273625746	-0.89580455\\
0.106966322024757	-0.289669633009248\\
0.0235950766583691	-0.594984356941613\\
0.0845451874151783	-0.806997019564963\\
0.104900662253661	-0.526750133142011\\
0.0594830483179512	-0.823813162153181\\
0.111940327014499	-0.479481649875851\\
0.0512962904368019	-0.807088831185495\\
0.110974030319992	-0.521534234332979\\
0.0577522856694644	-0.823566428367056\\
0.112227019021586	-0.479884502657897\\
0.0513129262891225	-0.807350708833395\\
0.111010150143497	-0.520884249218682\\
0.0576492376976389	-0.823413214603245\\
0.112223351400711	-0.480280052953319\\
0.0513728186251655	-0.807563973858598\\
0.111030757598379	-0.520361284196152\\
0.0575676268698583	-0.823286373091954\\
0.112219624895777	-0.480607862660636\\
0.051422558872083	-0.807739792856561\\
0.111047678569761	-0.519929800571638\\
0.0575003048971759	-0.823180164456706\\
0.112216316742346	-0.480882353983539\\
0.0514642375165078	-0.807886414586435\\
0.111061752594008	-0.519569725822114\\
0.0574441317900204	-0.823090458931686\\
0.112213395102049	-0.481114192832599\\
0.0514994593013675	-0.808009827604124\\
0.111073572685886	-0.519266474422042\\
};
\end{axis}
%
%
\begin{axis}[
width=5cm,
height=5cm,
at={(2.65in,0.331in)},
view={0}{90},
colormap/viridis,
xmin=-2,
xmax=2,
xlabel style={at = {(0.5,0)}},
xlabel={$\vartheta^{(1)}$},
ymin=-1,
ymax=1,
ylabel style={at = {(0,0.5)}},
ylabel={$\vartheta^{(2)}$},
yticklabels={,,},
xticklabels={,,},
title = {Port--Hamiltonian Optimizer},
]
 \addplot3[contour filled={number = 50,labels={false}},mesh/rows=50,mesh/cols=50,mesh/check=false,forget plot, patch type=bilinear
 ] table {H.dat};
\addplot [color=orange, line width=3pt]
  table[row sep=crcr]{%
1.9	1\\
1.89999908621382	0.999999146203232\\
1.89999634574791	0.999996585646132\\
1.89999177994975	0.999992319585279\\
1.89998539017821	0.999986349286269\\
1.89992613005681	0.999930979129558\\
1.8998214809747	0.999833198038975\\
1.8996716255243	0.999693174299225\\
1.89947675323871	0.999511081694832\\
1.89783419084367	0.997976001697306\\
1.89509857820368	0.995418394802826\\
1.89130097022727	0.991865873432015\\
1.88647599508352	0.987348955041771\\
1.86900749497567	0.970965954800983\\
1.84540136579028	0.948754533947611\\
1.81652608355849	0.921491956007284\\
1.78327039474016	0.889994677016268\\
1.73649313225686	0.845571687259977\\
1.6857190394663	0.797348039001021\\
1.63226428901274	0.746773067746493\\
1.57716842812712	0.695092133379318\\
1.5080088184612	0.631295136213572\\
1.43836740213243	0.568855318882871\\
1.36870093731967	0.508819085971922\\
1.29914489887713	0.451857667899015\\
1.20797627544001	0.382451962683775\\
1.11608075249029	0.319044370945756\\
1.02271007523783	0.261440623982038\\
0.927165334114407	0.209229053580064\\
0.795132426362742	0.146794390340538\\
0.658023794572816	0.0915836868932985\\
0.516984476691205	0.042381949921904\\
0.374698234745281	-0.00198595170672367\\
0.234935740627075	-0.0425632850706603\\
0.102732701058746	-0.079935584487616\\
-0.0172280071224691	-0.114648921752801\\
-0.121233885927335	-0.147236550118836\\
-0.196620002421501	-0.174211529660896\\
-0.256879073915717	-0.200092165794762\\
-0.301728690813634	-0.225069211763689\\
-0.331489071739927	-0.24932850721758\\
-0.348281435300383	-0.277268940631469\\
-0.346963924601245	-0.304653125264951\\
-0.329776517080097	-0.331669032647027\\
-0.299181127302743	-0.358446996103832\\
-0.257090727909214	-0.385460789246422\\
-0.206975492236153	-0.412317179067371\\
-0.151945079965006	-0.438951291178521\\
-0.0948713816719053	-0.46519206440883\\
-0.0428948327775945	-0.488754323819133\\
0.00648286357789724	-0.511500891208421\\
0.0515900094826211	-0.53311964415906\\
0.0912237351227792	-0.553264530149868\\
0.120122938176304	-0.569038094742835\\
0.144046490972093	-0.583241690211603\\
0.16283935223639	-0.595673400518031\\
0.176526413617999	-0.606174045766728\\
0.186327874474856	-0.615908653773451\\
0.189769018879384	-0.622701319126436\\
0.187504969239635	-0.626565467581752\\
0.180307146225369	-0.627651739520965\\
0.170259964939513	-0.626451279769171\\
0.157567376235483	-0.623426515935726\\
0.142924478373897	-0.618865557488853\\
0.126988670490448	-0.613076739784083\\
0.109675756234517	-0.606078068463564\\
0.0923379995460073	-0.598451199747082\\
0.0755924823726062	-0.590531573233015\\
0.0599386592320676	-0.582608767970448\\
0.0446303294195546	-0.574282088601846\\
0.0315221958219187	-0.56651829185148\\
0.0208973840622847	-0.559528166097405\\
0.0128724691349605	-0.553454497514785\\
0.00761252059856739	-0.548571548365942\\
0.00466162986985901	-0.544658676295795\\
0.00385457045430849	-0.541725833134023\\
0.00496227529670348	-0.539753348574028\\
0.00749482013453049	-0.538737411390349\\
0.0112127905427353	-0.538474638114025\\
0.0158494320269675	-0.538901080440744\\
0.0211461704734191	-0.539944107304326\\
0.028410834769471	-0.542022627947206\\
0.0358353785543005	-0.544779475156303\\
0.0429824666334652	-0.548028056255734\\
0.0495279641277516	-0.55158699059782\\
0.0547325228283026	-0.554929721322412\\
0.0590935534157801	-0.558250206596764\\
0.0625259730487866	-0.561430048031326\\
0.0650080452641057	-0.564370731790999\\
0.0665164887380692	-0.566889075958057\\
0.0672191423968561	-0.56905016199344\\
0.0671918620100267	-0.570810877759431\\
0.0665296638122927	-0.57214931984052\\
0.065106513120555	-0.573167108449149\\
0.0631610838550111	-0.573625172991193\\
0.0608817124569925	-0.573564090304827\\
0.0584357162303143	-0.573048333332431\\
0.0556748232650399	-0.572026131972753\\
0.0530911912697431	-0.570657940640856\\
0.0513296941487565	-0.564125538275932\\
};
\end{axis}
\end{tikzpicture}
    \vspace{-5mm}
    \caption{Comparison between the \textit{discrete} optimizer gradient descent and our \textit{continuous} port-Hamiltonian approach.}
    \label{fig:my_label}
    \vspace{-5mm}
\end{figure}
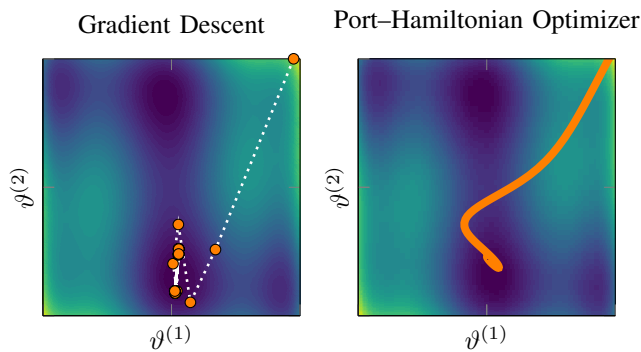
Port-Hamiltonian (PH) systems \cite{maschke1992port,duindam2009modeling,van2014port} have been introduced to model dynamical systems coming from different physical domains in a unified manner. This framework turned out to be fruitful in dealing with passivity based control (PBC) \cite{ortega2001putting,ortega2002interconnection,ortega2008control} since dissipativity information is explicitly encoded in PH systems, i.e. under mild assumptions those systems are passive. The aim of this work is to take advantage of such a structure and build a proper PH system associated to a neural network, in which the parameters of the latter are the states of the PH system. Within this framework, the weights evolve in time on a continuous trajectory along strictly decreasing level sets of the energy function, i.e. the objective function of the optimization problem, eventually landing in one of its minima. In this way, local optimality is intrinsically guaranteed by the PH dynamics.

This paper is structured as follows: Section \ref{sec:relwork} discusses previous works on continuous--time and energy--based approaches for neural networks. In Section \ref{sec:probset}, a formal introduction to neural networks to resolve some notational conflicts between control and learning theory. Section \ref{sec:PHNN} introduces port--Hamiltonian systems and their application to the training of neural networks.
Next, in Section \ref{sec:Experiments}, the performance of the proposed method is evaluated on a series of tasks and the results are discussed. Finally, in Section \ref{sec:concl} conclusions are drawn and future work is discussed.
\section{Related Work} \label{sec:relwork}
Recent works \cite{chen2018neural} have shown that it is possible to model residual layers as continuous blocks. This allows for a smooth transition between input and output: the input is integrated for a fixed time, which can be seen as the continuous analog of the number of network layers in the discrete case. By using the adjoint integration method Neural Ordinary Differential Equation Networks (ODE-Nets) offer improved memory efficiency and their performance is comparable to regular neural networks. ODE-Nets, however, are still optimized via discrete gradient descent methods. A similar idea was previously proposed in \cite{ruthotto2018deep}, which introduces Hamiltonian dynamics as a means of modeling network activations.  \cite{greydanus2019hamiltonian} introduces the Hamiltonian function as a useful physics-driven prior for learning conservative dynamics.
\cite{chaudhari2018deep} explores a connection between non-convex optimization and viscous Hamilton--Jacobi partial differential equations by introducing a modified version of stochastic gradient descent, Entropy--SGD. Entropy--SGD is applied to a function that is more convex in its input than the original loss function and yields faster convergence times. Similarly, a connection between Hamiltonian dynamics and learning was proposed in \cite{howse1996gradient} and \cite{sienko2004learning}. 
Energy-based models have been explored in the past \cite{ackley1985learning} \cite{hopfield1982neural}. Hopfield neural networks are designed to learn binary patters by iteratively reducing their \textit{energy} until convergence to an attractor. The energy is defined as a Lyapunov function of the weights of the network such that convergence to a local mininum is guaranteed. The binary-valued units $i$ of a Hopfield network are updated via a discrete procedure which checks if the weighted sum of the neighbouring units values does not reach a threshold, in which case the value of $i$ is flipped.
\section{Problem Setting} \label{sec:probset}
{%
\subsection{Notation} The set $\R$ ($\R^+$) is the the set of real (non negative real) numbers. The set of squared--integrable functions $z:\R\rightarrow\R^m$ is $\mathcal{L}_2^m$ while the set of $d$--times continuously differentiable functions is $\C^d$. Let $\langle\cdot,\cdot\rangle:\R^m\times\R^m\rightarrow\R$ denote the inner product on $\R^m$ and $\|v\|_2 \triangleq \sqrt{\langle v,v\rangle}$ its induced norm. 
The origin of $\R^n$ is $\mathbb{0}_n$. 
Let $\Ha:\R^n\rightarrow\R$ be $\mathcal{C}^1$ and let $\partial \Ha\in\R^n$ be its transposed gradient, i.e. $\partial\Ha\triangleq(\nabla\Ha)^\top\in\R^n$. In ambiguous cases, the variable with respect to which $\Ha$ is differentiated may appear as subscript, e.g. $\partial_x\Ha$. Indexes of vectors are indicated in superscripts, e.g. if $v$ is a vector $v^{(i)}$ indicates the $i$-th entry. Given two vectors $u,~v\in\R^n$, let $(u,v)\triangleq[u^\top,v^\top]^\top$.{\color{white}qqqqqqqqqqqqqqqqqqqqqqqqqqqqqqqqqqqqqq}
\subsection{Introduction to neural networks}
In order to provide a definition of \textit{neural networks} suitable for the scope of this paper, we must clarify the class of mathematical objects that they can handle. In particular, only networks whose input and outputs are vectors are treated. Indeed, the concepts presented hereafter can be naturally extended to more complex networks\footnote{Aforementioned networks often deal with multi--dimensional arrays (\textit{holors} \cite{moon2005theory}) which are referred as \textit{tensors} by the artificial intelligence community.}.
}

\begin{defn}[Neural Network]
	A neural network is a map 
	\begin{equation*}
	    f:\U\times\K\rightarrow\Y,
	\end{equation*}
	being $\U\subset\R^{n_u}$ the \textit{input space}, $\Y\subset\R^{n_y}$ the \textit{output space} and $\K\subseteq\R^{p}$ the manifold where the parameters characterizing the neural network live. The parameters, collected in a vector $\vartheta$, are assumed time dependent. Hence, 
	\begin{equation}\label{eq:neuralnet}
	    y = f(u,\vartheta(t))\quad u\in\U,~y\in\Y,~\vartheta\in\mathcal{K}.
	\end{equation}
\end{defn}

If samples $\hat{u}_i,\hat{y}_i$ $(i = 1,\dots,s)$ of the input and output spaces are provided, the parameters $\vartheta$ may be tuned in order to minimize an arbitrary cost function, e.g. the squared--norm of the output reconstruction error
\begin{equation}\label{eq:L2}
	\|e_i\|^2_2 \triangleq \|\hat{y}_i-f(\hat{u}_i,\vartheta)\|^2_2\quad\forall i = 1,\dots,s.
\end{equation} 
\begin{exmp}[Fully Connected Network]\label{ex:MLP}
	In a \textit{fully--connected} neural network, the $j$--th element $y_{i}^{(j)}$ of the output of the $i$--th layer is 
	\begin{equation}
		y_{i}^{(j)}  = \sigma\left(\sum_{k=1}^{h_{i-1}}y_{i-1}^{(j)}w_{i,j,k} + b_{i,j}\right)~~\forall j = 1,\dots,h_i~,
	\end{equation}
	where $h_{i}$ is the number of neurons in the $i$--th layer, $w_{i,j,k},b_{i,j}\in\R$, $\sigma_i:\R\rightarrow\R$ is called \textit{activation function}\footnote{Usually $\sigma_i$ is a nonlinear function.} and $y_0\triangleq u$. Indeed, $y_i$ can be symbolically rewritten in vector form as
	\begin{align*}
	y_i = 
	\sigma_i\left(
	W_iy_{i-1} +b_i
	\right),
	\end{align*}
	where 
	\begin{align*}
		W_i=\begin{bmatrix}
			w_{i,1,1}&w_{i,1,2}&\cdots&w_{i,1,h_{i-1}}\\
			w_{i,2,1}&w_{i,2,2}&\cdots&w_{i,2,h_{i-1}}\\
			\vdots&\vdots&\ddots&\vdots\\
			w_{i,h_i,1}&w_{i,h_i,2}&\cdots&w_{i,h_i,h_{i-1}}\\
			\end{bmatrix},
		b_i=\begin{bmatrix}
		b_{i,1}\\b_{i,2}\\\vdots\\b_{i,h_i}
		\end{bmatrix}
	\end{align*}
	and $\sigma_i$ is thought to be acting component--wise.
	Thus, for the $i$--th layer a vector $\vartheta_i$ containing all the weights and biases can be defined as
	\begin{equation}\label{eq:theta}
	    \vartheta_i=[w_{i,1,1},\dots,w_{i,h_i,h_{i-1}},b_{i,1},\dots,b_{i,h_i}]^\top\in\R^{h_i(1+ h_{i-1})}.
	\end{equation}
	Therefore, the overall vector containing all the parameters of a fully connected neural network with $l$ layers is
	\begin{equation}\label{eq:fc_par}
		\vartheta = \left(\vartheta_1,\vartheta_2,\dots,\vartheta_l\right)\in\R^p,
	\end{equation}
	where
	\begin{equation*}
		p = \sum_{i=1}^{l}h_i(1+ h_{i-1}).
	\end{equation*}
\end{exmp}  
%

\subsection{Training of a Neural Network}
Let $\U_s$, $\Y_s$ be finite and ordered subsets of the input and output spaces, i.e.

%
\begin{align*}
	\U_s &= \left\{\hat{u}_1,\hat{u}_2,\dots,\hat{u}_i,\dots,\hat{u}_s\right\}\subset\U,\nonumber\\
	\Y_s &= \left\{\hat{y}_1,\hat{y}_2,\dots,\hat{y}_i,\dots,\hat{y}_s\right\}\subset\Y,
\end{align*}
such that
\[
    \exists \Psi:\U\rightarrow\Y~:~\hat{y}_i=\Psi(\hat{u}_i)~\forall i=1,\dots,s~.
\]
The aim of the \textit{training} process of a neural network is to find a value of the parameters $\vartheta\in\K$ such that the elements of $\U_s$ are mapped by $f$ defined in (\ref{eq:neuralnet}) minimizing a given objective function dependent on the output samples, e.g. (\ref{eq:L2}). 

Let $\J:\U\times\Y\times\K\rightarrow\R$ be the objective (\textit{loss}) function; then, the solution of the training problem is 
\begin{equation}\label{eq:opti}
    \vartheta^* = \argmin\limits_{\vartheta} \J(\hat{u}_i,\hat{y}_i,\vartheta)\quad\forall \hat{u}_i\in\U_s,\hat{y}_i\in\Y_s.
\end{equation}
Consider a function $\Gamma:\U\times\Y\times\K\rightarrow\K$. Traditionally, a locally optimal solution of (\ref{eq:opti}) can be obtained by iterating a difference equation of the form:
\begin{equation}\label{eq:discrete_opti}
    \vartheta_{t+1} = \vartheta_t + \Gamma(\hat{u}_{t+1},\hat{y}_{t+1},\vartheta_t)\quad t = 1,\dots,s~,
\end{equation}
for a \textit{sufficient} \footnote{Here sufficient is intended in a statistical learning theory sense} number of steps, where the specific choice of $\Gamma$ determines the difference between training algorithms.

In contrast to such state--of--the--art methods, our approach is to equip the weights with continuous--time dynamics.
In particular, we model the behavior of the parameters with port-Hamiltonian systems. Due to the unique structure of this class of dynamical systems, asymptotic convergence toward an optimal solution will be automatically guaranteed.
\section{Training of Neural Networks: the Port--Hamiltonian Approach}\label{sec:PHNN}%
\subsection{Introduction to port--Hamiltonian systems}
A port--Hamiltonian (PH) system  has an input--state--output representation
\begin{equation}\label{eq:PH}
\left\{
\begin{matrix*}[l]
\dot{\xi} = 
[J(\xi)-R(\xi)]\partial\Ha(\xi) + g(\xi)v\\
z = g^\top(\xi)\partial \Ha(\xi)
\end{matrix*}
\right.~,
\end{equation}
%
with state $\xi\in\X\subset\R^n$, input $v\in\V\subset\R^m$ and output $z\in\mathcal{Z}\subset\R^{m}$. The function $\Ha:\X\rightarrow\R$ is called \textit{Hamiltonian function} and has the role of a generalized energy while $J(\xi)=-J^\top(\xi)\in\R^{n\times n}$ represents power preserving--interconnections, ${R}(\xi)=R(\xi)\geq 0$ models dissipative effects and ${g}(\xi)\in\R^{n\times m}$ describes the way in which external power is distributed into the system. In general, $\X$ is an $n$--dimensional manifold, $\mathcal{V}$ is a $m$--dimensional vector space and 
$\mathcal{Z} = \mathcal{V}^*$ is its dual space. Consequently the natural pairing $\langle v,z\rangle\triangleq z^\top v$ can be defined, which carries the unit measure of power (when modeling physical systems).
For compactness, from now on let us define $F(\xi)\triangleq J(\xi)-R(\xi)$ and omit the dependence on $\xi$ of $\Ha$ and $F$.
%
\begin{assum}
\begin{itemize}[Assumptions for PH systems]
    \item [1.] $F,g,\Ha$ are assumed smooth enough such that solutions are forward--complete for all initial conditions $\xi_0\in\X$, $v\in\mathcal{L}_2^m$;
    \item[2.] $\Ha$ is lower--bounded in $\X$, i.e.
    \[\exists \zeta\in\R:~\forall x\in\X~\Ha(\xi)>\zeta.\]
\end{itemize}
\end{assum}
From these assumptions it follows that PH systems are passive (see \cite{ortega2001putting,van2014port} ), i.e. 
\[\dot{\Ha}\leq z^\top v.\]
As a consequence, in the autonomous case ($v=0$) the Hamiltonian function is always non--increasing along trajectories. In particular,
\[\dot{\Ha} = \langle \partial\Ha,\dot{\xi}\rangle = -(\partial\Ha)^\top{R}\partial\Ha\leq 0~~\forall t\geq 0.\]
Thus, any strict minimum of $\Ha$ is a Lyapunov stable equilibrium point of the system. Furthermore, the control law $v = -kz$ $(k>0)$, usually referred as \textit{damping injection}, asymptotically stabilizes the equilibria \cite{ortega2001putting}.

Therefore, depending on the initial condition and the basins of attraction of the minima of $\Ha$, the state will eventually land in one minimum point of the Hamiltonian function. This latter property is the key that allows the use of PH systems for the training of neural networks.
%
\subsection{Equip the network with port--Hamiltonian dynamics}\
The proposed approach consists in describing the dynamics of the neural network's parameters using an autonomous PH system. In fact, if the Hamiltonian function coincides with the loss function of the learning problem, i.e. $\Ha\triangleq \J$, we guarantee asymptotic convergence to a minimum of $\J$, i.e. solution of the problem (\ref{eq:opti}) and hence successful training of the neural network.

Generally, a desirable property of the parameter dynamics is to reach a minimum of $\J$ with null velocity $\dot{\vartheta}$. In the port--Hamiltonian framework, this can be achieved with mechanical--like equations. Let $\omega\triangleq M(\vartheta)\dot{\vartheta}$ be a vector of fictitious generalized momenta where $M=M^\top>0$ is the generalized inertia matrix. The role of $M$ is to give different \textit{weight} to the parameters and model specific couplings between their dynamics. Then, let the state of the PH system be
\begin{equation*}
	\xi \triangleq (\vartheta,\omega)\in\R^{2p}. 
\end{equation*}
Hence, the loss function might be redefined adding a term $\J_{\tt kin}(\vartheta,\omega)$ equivalent to a pseudo kinetic energy:
\begin{equation*}
	\J^*(\hat{u},\hat{y},\xi) \triangleq \J(\hat{u},\hat{y},\vartheta) + \underbrace{\omega^\top M^{-1}(\vartheta)\omega
	}_{\J_{\tt kin}(\vartheta,\omega)}.
\end{equation*}
Note that $\J(\hat{u},\hat{y},\vartheta)$ represents the potential energy of the fictitious mechanical system.

As for general $p$ degrees--of--freedom mechanical system in PH form, the choice of $J$ and $R$ is:
\[
    J\triangleq\begin{bmatrix}O_p&I_p\\-I_p&O_p\end{bmatrix}\in\R^{2p\times2p},~R\triangleq\begin{bmatrix}O_p&O_p\\O_p&B\end{bmatrix}\in\R^{2p\times2p},
\]
where $O_p$, $I_p$ are respectively the $p$--dimensional zero and identity matrices while $B=B^\top>0$, $B\in\R^{p\times p}$.
Therefore, the autonomous PH model of the parameters dynamics obtained by setting $\Ha=\J^*$ is:
\begin{align}\label{eq:PH_weights}
    &\begin{bmatrix}\dot{\vartheta}\\\dot{\omega}\end{bmatrix} =\left(J-R\right)\begin{bmatrix}\partial_{\vartheta}\J^*\\\partial_{\omega}\J^*\end{bmatrix}\nonumber\\
    \Leftrightarrow  & ~~~\dot{\xi} = \underbrace{\begin{bmatrix}0&I_n\\-I_n&-B\end{bmatrix}}_{{F}}\partial{\J^*}.
 \end{align}
Hence, trajectories of $\vartheta,\omega$ will unfold on continuously decreasing level sets of $\J^*$ which plays the role of a \textit{generalized} energy.
\begin{exmp}\label{exmp:L2pot}
	Suppose
	\begin{equation*}
		M(\vartheta) = I_p\Rightarrow\omega = \dot{\vartheta}
	\end{equation*}
	and let $\J$ be the mean--squared--error loss. Therefore, a possible choice of the loss function $\J^*$ is
	\begin{equation}\label{eq:loss}
	\mathcal{J}^*(\hat{u},\hat{y},\xi) = \frac{1}{2}\left[\alpha\|\hat{y}-f(\hat{u},\vartheta(t))\|_2^2 + \beta{\vartheta}^\top\vartheta + \dot{\vartheta}^\top\dot{\vartheta}\right],
	\end{equation}
	with $\alpha,\beta\in\R^+$. Indeed, every minima of $\mathcal{J}^*$ is placed in $\dot{\vartheta} = \mathbb{0}_p$. The gradient of $\J^*$ is 
	\begin{equation*}
		\partial\J^* = \left(\alpha\frac{\partial f}{\partial \vartheta}[\hat{y}-f(\hat{u},\vartheta)] + \beta\vartheta,\dot{\vartheta}\right).
	\end{equation*}
	With this choice of $\J^*$, the dynamics of the parameters become a (nonlinear) second--order ordinary differential equation:
	\begin{equation*}
		\ddot{\vartheta} + \alpha\frac{\partial f}{\partial \vartheta}[\hat{y}-f(\hat{u},\vartheta)] + \beta\vartheta + B\dot{\vartheta} = \mathbb{0}_p ~.
	\end{equation*}
\end{exmp}
\vspace{3mm}
\begin{rem}
    	{
    	The term $\beta{\vartheta}^\top\vartheta$ in (\ref{eq:loss}) is introduced as a \textit{regularization} tool. Regularization is a fundamental technique in machine learning that is widely used in order to find solutions with smaller norm (e.g. \textit{weight decay}) or to enforce sparsity in the parameters (e.g. \textit{L1--regularization}).
    	Note that this is a particular case of the Tikhonov regularization term $\|\Lambda\vartheta\|_2^2$ with $\Lambda = \beta I_p$} \cite{golub1999tikhonov}, also known as weight decay \cite{krogh1992simple}.
\end{rem}
\begin{rem}
    In the context of mechanical--like PH system, a consistent choice of the power port is 
    \[
        g \triangleq 
            \begin{bmatrix}
                O_p&O_p\\O_p&I_p
            \end{bmatrix},
    \]
    selecting as input the fictitious generalized forces and as output the velocities $z = \dot{\vartheta}$.
    Hence, during the training of the neural network, a control input $v = -k(t)\dot{\vartheta}$ ($k(t)\geq 0~\forall t$) might be applied to dynamically modify the rate with which the parameters of the network are optimized. This opens different scenarios for designing a $k(t)$ which increases the probability of reaching the global minimum of $\J^*$. In fact, the choice of $k$ determines the shape of the basins of attraction of the minima of $\J^*$ \cite{massaroli2019multistable}. This open problem is left for future work.
\end{rem}
\vspace{5mm}
\begin{defn}[Port--Hamiltonian neural network]
	We define a \textit{port--Hamiltonian neural network} (PHNN) as a neural network whose parameters $\vartheta$ have continuous--time dynamics (\ref{eq:PH_weights}):
	\begin{equation}\label{eq:HDNN}
	    \left\{
	        \begin{matrix*}[l]
	            \dot{\xi} = {F}\partial\J^*\\
	            y = f(u,\xi)
	        \end{matrix*}
	    \right. ~.
	\end{equation}
	\vspace{1mm}
\end{defn}
Note that a PHNN is uniquely defined by the triplet ($f,F,\J^*$).
\begin{exmp}[Linear Classifier]\label{exmp:linclass}
Consider a fully connected network (see Example \ref{ex:MLP}) with a single layer, $h$ neurons\footnote{In this case, the number of neurons equals the dimension of the input space.} and $l$ classes, i.e. $u \in\U\subset\R^{h},~y \in\Y\subset\R^{l}$. Therefore,
\begin{align}
y = f(u,\vartheta)\triangleq
\begin{bmatrix}
w_{1,1}&w_{1,2}&\cdots&w_{1,h}&b_1\\
w_{2,1}&w_{2,2}&\cdots&w_{2,h}&b_2\\
\vdots&\vdots&\ddots&\vdots&\vdots\\
w_{l,1}&w_{l,2}&\cdots&w_{l,h}&b_l
\end{bmatrix}
\begin{bmatrix}
u\\1
\end{bmatrix}.
\end{align}
Let $\vartheta$ and the loss function $\J^*$ be defined as in (\ref{eq:theta}) and (\ref{eq:loss}) respectively. Hence,
\begin{align}
    \xi = (\vartheta,\dot{\vartheta})\in\R^{2l(h+1)}.
\end{align}
Then,
\begin{equation*}
    \partial\J^* = 
        \begin{bmatrix}
            \partial_{\vartheta}\J^*\\
            \partial_{\dot{\vartheta}}\J^*
        \end{bmatrix} = 
        \begin{bmatrix}
            \alpha\langle(\hat{u},1),\hat{y}-f(\hat{u},\vartheta)\rangle + \beta \vartheta\\
            \dot{\vartheta}	
        \end{bmatrix},
\end{equation*}
\begin{align*}
    \dot{\xi} \nonumber &={F}\partial\J^*= 
        \begin{bmatrix}
            \dot{\vartheta}\\
-           \alpha\langle(\hat{u},1),\hat{y}-f(\hat{u},\vartheta)\rangle - \beta \vartheta - B\dot{\vartheta}
        \end{bmatrix}.
\end{align*}
\end{exmp}
\vspace{0.1cm}
\subsection{Training of PHNNs}
Let us assume that a dataset of inputs $\U_s$ and outputs (\textit{labels}) $\Y_s$ is available. In this section two training techniques will be introduced.
\subsubsection{Sequential data training} 
As already pointed out, given an initial condition $\xi_0$, the system will converge to a minimum $\vartheta^*$ of $\J$, i.e. to a minimum $(\vartheta^*,\mathbb{0}_p)$ of $\J^*$. However, the location of the minima strictly depends on the training data.
The \textit{sequential} training approach relies on iteratively feeding one tuple $\hat{u}_i,\hat{y}_i$ to the PHNN integrating the differential equation for a time $t^*$ in each iteration. This process can be carried out from scratch several times (\textit{epochs}).
 
Let $\tau$ be a \textit{timer}, i.e. $\dot{\tau} = 1$ and $\zeta$ a \textit{cycle counter}, both initialized to 0. 
After the initialization step, a first tuple $\hat{u}_1,\hat{y}_1$ is fed to the PHNN and integration starts from $\tau = 0$. When $\tau = t^*$ a new tuple is fetched, $\tau$ is reset, $\zeta$ is increased by 1 and the state $\xi$ is carried over. The process is repeated until $\zeta = s$ and the first $epoch$ is complete, at which point the first tuple will be fetched once again. This technique is reminiscent of the way in which SGD updates are performed in practice. 

\begin{figure}[t]
	\centering
	\begin{tikzpicture}[scale=0.95, every node/.style={scale=0.95}]
	\small
	\fill[gray!20, draw = black,thick] (0,0) ellipse (2cm and 1.8cm);
	\draw (0,-0.1) node[align = center](flow)  {
		$
		\begin{matrix*}
		\begin{bmatrix}
		\dot{\xi}\\
		\dot{\tau}\\
		\dot{\zeta}\\
		\dot{\hat{u}}\\
		\dot{\hat{y}}
		\end{bmatrix} = 
		\begin{bmatrix}
		{F}\partial\J^*(\hat{u},\hat{y},\xi)\\
		1\\
		0\\
		\mathbb{0}_{n_u}\\
		\mathbb{0}_{n_y}		
		\end{bmatrix}
		\\
		y = f(\hat{u},\xi)
		\end{matrix*}
		$};
	\draw [thick,-latex] plot [smooth, tension=1.2] coordinates { (0.7,1.705) (1.8,1.5) (1.89,0.51)};
	\draw [thick,-latex] plot [smooth, tension=1.2] coordinates {(-0.7,-1.705) (-1.8,-1.5) (-1.89,-0.51)};
	\draw (.5,1.95) node {$\tau = t^*\land \zeta\neq s$};
	\draw (.1,-2) node {$\tau = t^*\land \zeta = s$};
	\draw (3.3,0) node[align = center](jump) {
		$\begin{bmatrix}
			\xi^+\\
			\tau^+\\
			\zeta^+\\
			\hat{u}^+\\
			\hat{y}^+
		\end{bmatrix} = 
		\begin{bmatrix}
			\xi\\
			0\\
			\zeta+1\\
			\hat{u}_{\zeta+1}\\
			\hat{y}_{\zeta+1}
		\end{bmatrix}$
		};
    \draw (-3.1,0) node[align = center](jump2) {
		$\begin{bmatrix}
			\xi^+\\
			\tau^+\\
			\zeta^+\\
			\hat{u}^+\\
			\hat{y}^+
		\end{bmatrix} = 
		\begin{bmatrix}
			\xi\\
			0\\
			0\\
			\hat{u}_{\zeta+1}\\
			\hat{y}_{\zeta+1}
		\end{bmatrix}$
		};
%
	%
	\end{tikzpicture}
	\caption{\footnotesize Hybrid automata: Conceptual representation of the hybrid system modeling the \textit{sequential training} of the neural network.}
	\label{fig:HA}
	\vspace{-5mm}
\end{figure}
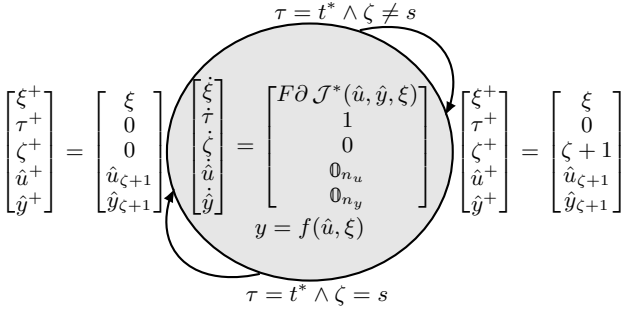

The PHNN with the \textit{update and converge} training can be represented by means of an \textit{hybrid dynamical system} (see \cite{van2000introduction}) whose graphical representation is shown in Fig. \ref{fig:HA}.
\subsubsection{Batch training}
In the \textit{batch} method the neural network is trained using the entire dataset simultaneously. 
To do this, the loss function is redefined as the average of the losses of each sample, i.e. 
\begin{equation*}
\mathcal{J}^*_{\tt batch}(\U_s,\Y_s,\xi) \triangleq \frac{1}{s}\sum_{i=1}^s\mathcal{J}^*(\hat{u}_i,\hat{y}_i,\xi)\triangleq\sum_{i=1}^s\mathcal{J}^*_i.
\end{equation*}
%
Thanks to the linearity of differentiation, it is also possible to compute the gradient as the average of the gradients of the single losses:
\begin{equation*}
\partial\mathcal{J}^*_{\tt batch} = \frac{1}{s}\sum_{i=1}^s\partial\mathcal{J}^*_i.
\end{equation*} 
Then, the training is simply achieved by integrating 
\begin{equation*}
	\dot{\xi} = F\partial\J^*_{\tt batch}.
\end{equation*}
\begin{rem}
Note that the sequential training will stop at one of the minima of $\J^*(\hat{u}_\zeta,\hat{y}_\zeta,\xi)$ (depending of the time at which the procedure is stopped), which might not necessarily coincide with a minimum of $\J_{\tt batch}^*(\U_s,\Y_s,\xi)$. 
\end{rem}
\subsection{Computational complexity}
Theoretical space and time complexity of the proposed method are comparable to standard gradient descent and depend on the specific ODE solving algorithm employed. Let $p$ be the number of parameters of the neural network to optimize. Regular gradient descent has space complexity linear in $p$ (i.e $\mathcal{O}(p)$), whereas PHNNs require an additional state per parameter, the momentum, thus also yielding linear space complexity. Similarly, time complexity is linear in $\epsilon$, the number of gradient descent steps necessary for convergence to a neighbourhood of a minimum. 

\section{Numerical Experiment}\label{sec:Experiments}
The effectiveness of PHNNs has been evaluated on the following two classes of numerical experiments. As an initial test, the PHNN has been tasked with learning a linear boundary between two classes of points by using a sequential training approach. The second experiment, on the other hand, deals with non-linear vector field approximation via the use of the batch training method. All the experiments have been implemented in \textit{Python}\footnote{The code is available at: \url{https://github.com/Zymrael/PortHamiltonianNN}.} 

\subsection{Task 1: Learning a linear boundary}
\begin{figure}[!t]
	\centering
	\input{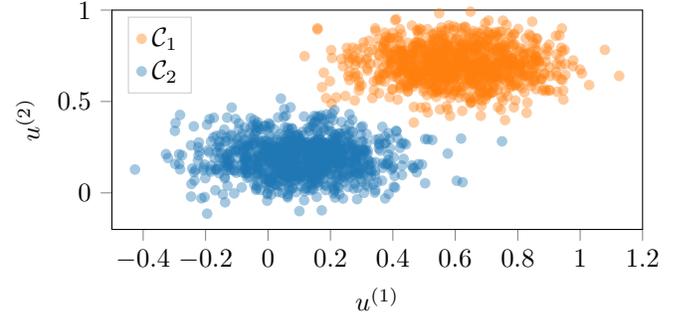}
	\vspace{-6mm}
	\caption{\footnotesize Dataset used to train the neural network in the numerical experiment}
	\label{fig:dataset}
\end{figure}

\begin{figure}[!t]
	\centering
	\input{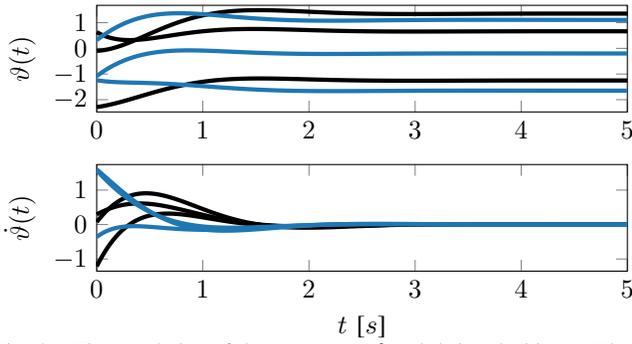}
	\vspace{-9mm}
	\caption{\footnotesize Time evolution of the parameters $\vartheta$ and their velocities $\omega$. Black indicates parameters of $y^{(1)}$ while blue parameters of $y^{(2)}$.}
	\label{fig:weights}
\end{figure}

\begin{figure}[!]
	\centering
	\input{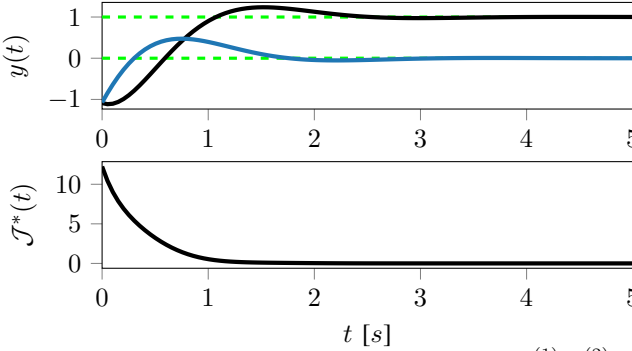}
	\vspace{-9mm}
	\caption{\footnotesize [Above] Time evolution of the estimated output $y$. $y^{(1)}$, $y^{(2)}$ are indicated with black and blue lines respectively. [Below] Decay in time of the loss function $\mathcal{J}^*$.}
	\label{fig:out_loss}
\end{figure}

Consider the PHNN of Example \ref{exmp:linclass} in the case $h = 2,~l = 2$, i.e., $u\triangleq[u^{(1)},u^{(2)}]^\top\in\R^2,~y\triangleq[y^{(1)},y^{(2)}]^\top\in\R^2$. The aim of the numerical experiment is to learn a linear boundary separating two classes of points sampled from two bivariate Gaussian distributions $\mathcal{N}_1$, $\mathcal{N}_2$. We will refer to the two classes as $\C_1$ and $\C_2$.
The neural network model is
\[\begin{bmatrix}y^{(1)}\\y^{(2)}\end{bmatrix} =\begin{bmatrix}w_{1,1}u^{(1)} + w_{1,2}u^{(2)} + b_1\\w_{2,1}u^{(1)} + w_{2,2}u^{(2)} + b_2\end{bmatrix}\]
and the corresponding parameter vector is
\[\vartheta\triangleq[w_{1,1},w_{1,2},w_{2,1},w_{2,2},b_1,b_2]^\top\in\R^6\Rightarrow\xi\in\R^{12}.\]
The dataset has been built sampling a total of 1000 points from each distribution and has been collected in $\U_s$ in a shuffled order. The result is shown in Fig. \ref{fig:dataset}.
%
%
The corresponding reference outputs have been computed and stored in $\Y_s$. In particular,
\begin{equation}
    \hat{y} = \Psi(\hat{u})\triangleq\left\{
        \begin{matrix*}[l]
            [1,0]^\top~~\hat{u}\in\C_1\\
            [0,1]^\top~~\hat{u}\in\C_2\\
        \end{matrix*}
    \right.\quad\forall \hat{u} \in \U_s.
\end{equation}
Then, the training procedure has been performed on a single sample $(\hat{u},\hat{y}) = ([0.6,0.6]^\top,[1,0]^\top)$. The weights and their velocities have been initialized as $\xi_0 = [0.6,-2.3,-0.1,-1.1,-1.2,0.3,$ $-1.2,0.3,0.2,1.6,-0.4,1.6]^\top$. The system parameters have been chosen as, $B = I_6$, $\alpha=1$ and $\beta=0$. The resulting ODE has been numerically integrated for a time $t_f = 5s$.
The resulting weight trajectories are reported in Fig. \ref{fig:weights}. Black is used to highlight parameters that are used to compute the first output element $y^{(1)}$ whereas blue is similarly used for parameters of $y^{(2)}$. Furthermore, the time evolution of the output of the neural network and the one of the loss function are shown in Fig. \ref{fig:out_loss}. 


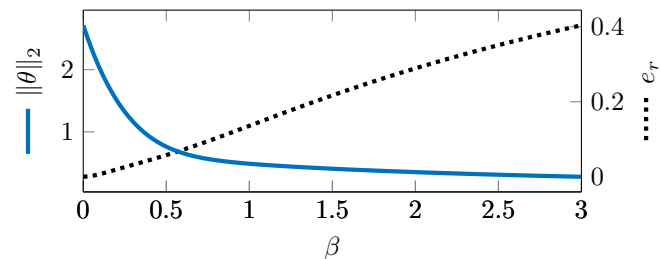
\begin{figure}[!b]
	\centering
%
%
\definecolor{mycolor1}{rgb}{0.85098,0.32549,0.09804}%
\definecolor{mycolor2}{rgb}{0.00000,0.44706,0.74118}%
%

\begin{tikzpicture}
\pgfplotsset{set layers}
\begin{axis}
[
width=0.95*\columnwidth,
height=4cm,
%
xmin=0,xmax=3,
domain=0:15,
axis y line*=right,
xlabel={$\beta$},
ylabel style = {align=center},
ylabel={\ref{err} $e_r$},
legend style={draw=none},
legend style={at={(.3,0.8)}}
]
\addplot [ultra thick, color=black, dotted]
table[row sep=crcr]{%
	0	3.29727809797545e-06 \\
	0.0303030303030303	0.000919482366455613 \\
	0.0606060606060606	0.00312631399193632 \\
	0.0909090909090909	0.00612705030462364 \\
	0.121212121212121	0.00958634586132847 \\
	0.151515151515152	0.0132859759111303 \\
	0.181818181818182	0.017091243239988 \\
	0.212121212121212	0.0209257335382162 \\
	0.242424242424242	0.0247525526219097 \\
	0.272727272727273	0.028560563370012 \\
	0.303030303030303	0.0323544509544393 \\
	0.333333333333333	0.0361476919676284 \\
	0.363636363636364	0.0399577119993677 \\
	0.393939393939394	0.0438026600696098 \\
	0.424242424242424	0.0476993756174516 \\
	0.454545454545455	0.0516622015374567 \\
	0.484848484848485	0.0557024155118805 \\
	0.515151515151515	0.0598280655283917 \\
	0.545454545454545	0.064044060318148 \\
	0.575757575757576	0.0683524505353423 \\
	0.606060606060606	0.0727527889676343 \\
	0.636363636363636	0.0772425309571249 \\
	0.666666666666667	0.0818174400217852 \\
	0.696969696969697	0.086471973032248 \\
	0.727272727272727	0.0911996319665787 \\
	0.757575757575758	0.095993272588865 \\
	0.787878787878788	0.100845372864134 \\
	0.818181818181818	0.105748253057295 \\
	0.848484848484849	0.110694256499795 \\
	0.878787878787879	0.115675892009016 \\
	0.909090909090909	0.120685942286662 \\
	0.939393939393939	0.125717543296482 \\
	0.96969696969697	0.130764238411743 \\
	1	0.135820012595611 \\
	1.03030303030303	0.140879309219428 \\
	1.06060606060606	0.145937030543629 \\
	1.09090909090909	0.150988539018549 \\
	1.12121212121212	0.156029633414713 \\
	1.15151515151515	0.161056531677703 \\
	1.18181818181818	0.166065848991955 \\
	1.21212121212121	0.17105456985451 \\
	1.24242424242424	0.176020020583065 \\
	1.27272727272727	0.180959841529506 \\
	1.3030303030303	0.185871959625001 \\
	1.33333333333333	0.19075456262562 \\
	1.36363636363636	0.195606073188405 \\
	1.39393939393939	0.200425125234006 \\
	1.42424242424242	0.20521054335876 \\
	1.45454545454545	0.209961324861825 \\
	1.48484848484848	0.214676606447178 \\
	1.51515151515152	0.219355681740345 \\
	1.54545454545455	0.223997939799726 \\
	1.57575757575758	0.228602892756058 \\
	1.60606060606061	0.233170140703854 \\
	1.63636363636364	0.237699372761274 \\
	1.66666666666667	0.242190349890465 \\
	1.6969696969697	0.246642891293022 \\
	1.72727272727273	0.251056885920995 \\
	1.75757575757576	0.255432270567941 \\
	1.78787878787879	0.259769028423652 \\
	1.81818181818182	0.264067182963208 \\
	1.84848484848485	0.268326795346735 \\
	1.87878787878788	0.272547958497306 \\
	1.90909090909091	0.276730795884557 \\
	1.93939393939394	0.28087545563354 \\
	1.96969696969697	0.284982109861598 \\
	2	0.289050951237746 \\
	2.03030303030303	0.293082191052906 \\
	2.06060606060606	0.297076057984927 \\
	2.09090909090909	0.301032793730698 \\
	2.12121212121212	0.304952655891271 \\
	2.15151515151515	0.308835911740281 \\
	2.18181818181818	0.312682841846243 \\
	2.21212121212121	0.316493730509718 \\
	2.24242424242424	0.320268874447055 \\
	2.27272727272727	0.324008579472672 \\
	2.3030303030303	0.327713146636499 \\
	2.33333333333333	0.331382905343222 \\
	2.36363636363636	0.335018167644085 \\
	2.39393939393939	0.338619270237271 \\
	2.42424242424242	0.342186514757892 \\
	2.45454545454545	0.345720235372864 \\
	2.48484848484848	0.349220788101624 \\
	2.51515151515152	0.352688499527639 \\
	2.54545454545455	0.356123695262475 \\
	2.57575757575758	0.359526715501796 \\
	2.60606060606061	0.362897897282665 \\
	2.63636363636364	0.366237577062165 \\
	2.66666666666667	0.369546098603299 \\
	2.6969696969697	0.372823788617522 \\
	2.72727272727273	0.376070982384049 \\
	2.75757575757576	0.37928801909905 \\
	2.78787878787879	0.382475229383591 \\
	2.81818181818182	0.38563294520474 \\
	2.84848484848485	0.388761496832078 \\
	2.87878787878788	0.391861208321821 \\
	2.90909090909091	0.394932411345604 \\
	2.93939393939394	0.397975429336226 \\
	2.96969696969697	0.400990574623142 \\
	3	0.403978169760702 \\
};
\label{err}
\end{axis}
\begin{axis}
[
width=0.95*\columnwidth,
height=4cm,
xmin=0,xmax=3,
axis y line*=left,
axis x line*=bottom,
ylabel = {\ref{theta2} $\|\theta\|_2$},
legend style={draw=none},
legend style={at={(0.355,0.95)}}
]
\addplot [color=mycolor2, ultra thick]
table[row sep=crcr]{%
	0	2.71387208642868 \\
	0.0303030303030303	2.47770387672656 \\
	0.0606060606060606	2.26352075535896 \\
	0.0909090909090909	2.06953195875802 \\
	0.121212121212121	1.89408961128625 \\
	0.151515151515152	1.73567636450804 \\
	0.181818181818182	1.59289364257306 \\
	0.212121212121212	1.46445044801033 \\
	0.242424242424242	1.34915273441695 \\
	0.272727272727273	1.24589339002278 \\
	0.303030303030303	1.15364292233432 \\
	0.333333333333333	1.07144098217216 \\
	0.363636363636364	0.998388887824748 \\
	0.393939393939394	0.933643338739021 \\
	0.424242424242424	0.876411456696024 \\
	0.454545454545455	0.825947248947958 \\
	0.484848484848485	0.781549440485918 \\
	0.515151515151515	0.742560563165119 \\
	0.545454545454545	0.708366985201568 \\
	0.575757575757576	0.678399522431765 \\
	0.606060606060606	0.652134226227774 \\
	0.636363636363636	0.629092951291654 \\
	0.666666666666667	0.608843410802438 \\
	0.696969696969697	0.590998525240305 \\
	0.727272727272727	0.575215002176104 \\
	0.757575757575758	0.561191204422701 \\
	0.787878787878788	0.548664398081362 \\
	0.818181818181818	0.537407599733835 \\
	0.848484848484849	0.527226154696658 \\
	0.878787878787879	0.51795424789457 \\
	0.909090909090909	0.509451459484307 \\
	0.939393939393939	0.501599492910755 \\
	0.96969696969697	0.494299135395369 \\
	1	0.487467498885848 \\
	1.03030303030303	0.481035556154582 \\
	1.06060606060606	0.474945961608612 \\
	1.09090909090909	0.469151177327451 \\
	1.12121212121212	0.463611861146778 \\
	1.15151515151515	0.458295464513751 \\
	1.18181818181818	0.453175076111187 \\
	1.21212121212121	0.448228432771541 \\
	1.24242424242424	0.44343709737185 \\
	1.27272727272727	0.438785773232463 \\
	1.3030303030303	0.434261736390009 \\
	1.33333333333333	0.429854365511266 \\
	1.36363636363636	0.425554756745185 \\
	1.39393939393939	0.421355407053332 \\
	1.42424242424242	0.417249953023734 \\
	1.45454545454545	0.413232958593577 \\
	1.48484848484848	0.409299751597146 \\
	1.51515151515152	0.405446262981263 \\
	1.54545454545455	0.401668942555681 \\
	1.57575757575758	0.397964640383645 \\
	1.60606060606061	0.394330547954254 \\
	1.63636363636364	0.390764130664142 \\
	1.66666666666667	0.387263084490421 \\
	1.6969696969697	0.383825301445204 \\
	1.72727272727273	0.380448824662165 \\
	1.75757575757576	0.377131834467342 \\
	1.78787878787879	0.373872625181117 \\
	1.81818181818182	0.370669589657018 \\
	1.84848484848485	0.367521204891116 \\
	1.87878787878788	0.364426022945148 \\
	1.90909090909091	0.361382660789018 \\
	1.93939393939394	0.358389795389259 \\
	1.96969696969697	0.355446156554957 \\
	2	0.352550522994489 \\
	2.03030303030303	0.349701718268883 \\
	2.06060606060606	0.346898607051175 \\
	2.09090909090909	0.344140094217209 \\
	2.12121212121212	0.34142511938723 \\
	2.15151515151515	0.338752658216446 \\
	2.18181818181818	0.336121717232935 \\
	2.21212121212121	0.333531337988862 \\
	2.24242424242424	0.330980588941404 \\
	2.27272727272727	0.328468565813035 \\
	2.3030303030303	0.32599439907107 \\
	2.33333333333333	0.323557230140057 \\
	2.36363636363636	0.321156240406398 \\
	2.39393939393939	0.318790621882022 \\
	2.42424242424242	0.316459615156561 \\
	2.45454545454545	0.314162463675467 \\
	2.48484848484848	0.311898419969546 \\
	2.51515151515152	0.309666777790996 \\
	2.54545454545455	0.307466852391422 \\
	2.57575757575758	0.305297969624307 \\
	2.60606060606061	0.303159476459054 \\
	2.63636363636364	0.301050738425639 \\
	2.66666666666667	0.29897113382958 \\
	2.6969696969697	0.296920068278095 \\
	2.72727272727273	0.294896957779564 \\
	2.75757575757576	0.292901231044548 \\
	2.78787878787879	0.290932336578809 \\
	2.81818181818182	0.288989735859558 \\
	2.84848484848485	0.287072904626973 \\
	2.87878787878788	0.285181335058054 \\
	2.90909090909091	0.283314526778068 \\
	2.93939393939394	0.281471996035045 \\
	2.96969696969697	0.279653276857083 \\
	3	0.277857907690036 \\
};
\label{theta2}
\end{axis}
\end{tikzpicture}
	\vspace{-9mm}
	\caption{\footnotesize Effect of the regularization term on the output reconstruction error and on the parameters vector norm.}
	\label{fig:reg}
\end{figure}

In order to show the effect of the regularization term  $\beta\vartheta^\top\vartheta$ we performed the same experiment multiple times varying $\beta$ in the interval $[0,3]$. At each iteration, the relative output tracking squared error 
\[
    e_{r} \triangleq \frac{\|\hat{y}-y(t_f)\|_2^2}{\|\hat{y}\|_2^2}
\]
and the norm $\|\vartheta\|_2$ of the parameters vector, have been computed. This shows that the effect of $\beta$ is comparable to the effect of weight decay in neural networks optimized via discrete methods, namely a reduction of the parameter norm. The results are shown in Fig. \ref{fig:reg}.

\begin{figure}[!t]
	\centering
	\input{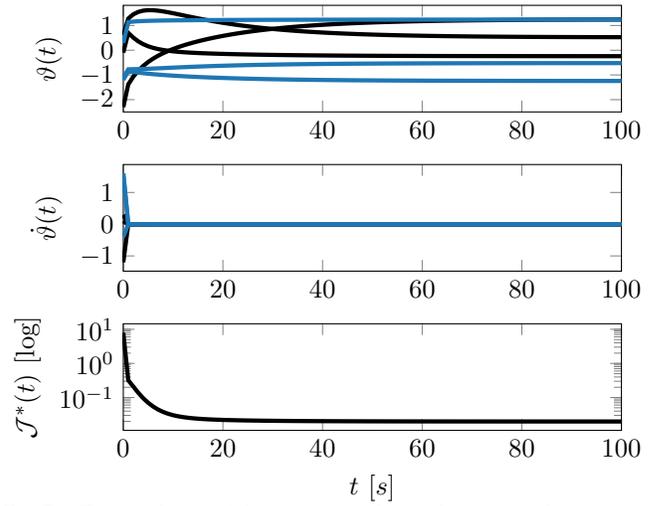}
	\vspace{-9mm}
	\caption{\footnotesize Time evolution of the parameters during the sequential training on the linear boundary problem. Black indicates parameters of $y^{(1)}$ while blue parameters of $y^{(2)}$.}
	\label{fig:SecTraj}
\end{figure}

\begin{figure}[!b]
	\centering
	\input{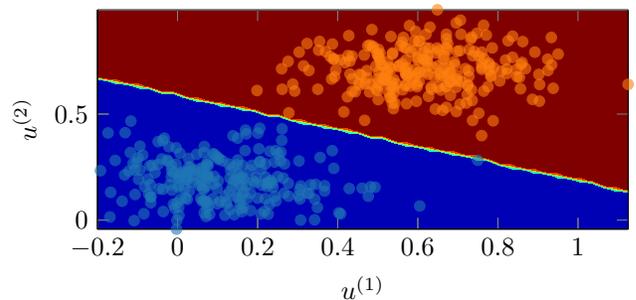}
	\vspace{-5mm}
	\caption{\footnotesize Decision boundary plot and test set.}
	\label{fig:decbound}
\end{figure}

Subsequently, the dataset $\U_s,~\Y_s$ has been split in a \textit{training set} and a \textit{test set} with a ratio of $3:1$. Then, the optimization of the network's parameters has been performed with the \textit{sequential} method by using exclusively training set data while classification accuracy of the trained network has been evaluated on the test set. The chosen values of the parameters are the following: $B = 100I_6$, $\alpha = 1$, $\beta = 0.001$, $t^*=0.1$ and $\xi_0$ has been initialized as before. The training procedure has been carried out for 100 epochs. The time evolution of the parameters and the loss function (one value per epoch) is shown in Fig. \ref{fig:SecTraj}. As the loss is non-increasing with the number of epochs, the parameters converge to constant values. Furthermore, a \textit{decision boundary} has been plotted in Fig. \ref{fig:decbound} which shows how the the linear boundary learned by the network during training correctly separates the two classes, correctly classify all the points of the test set.

\subsection{Task 2: Learning a vector field}
To further test the performance of the proposed training approach in a more complex scenario, the problem of approximating a vector field has been addressed. Consider a nonlinear ODE
\begin{equation}\label{nl_oscill}
    \frac{d u}{d x} = \Phi(u)~~~~u\in\R^n,~\Phi:\R^n\rightarrow\R^n,~x\in\R.
\end{equation}
The learning task consisted in training a fully--connected neural network to approximate the vector field $\Phi$ by using only some samples of the state $u$. 
Thus, input data have been generated collecting state observations along a trajectory in $s+1$ points $x_i$
\[\hat{u}_i \triangleq u(x_i).\]
The corresponding labels have been computed approximating the state derivative via forward difference, i.e.
\[
    \hat{y}_i = \frac{u(x_{i+1})-u(x_i)}{x_{i+1}-x_i}\approx\Phi(u(x_i)) ~~\forall i=1,\dots,s~.
\]
Hence, the input and output datasets $\U_s$, $\Y_s$ have been built and, then, the neural network has been trained with the PHNN method. The objective was to obtain a network able to infer the knowledge of the vector field, learned on a single trajectory, to a wider region of the state space. Thus, the metric chosen to evaluate the training performance has been the absolute approximation error in a domain $\D$:
\[
    \mathcal{E}(u)\triangleq\|\Phi(u)-f(u,\vartheta^*)\|_2\quad u\in\D,
\]
where $\vartheta^*$ is the optimized vector of parameters. Notice that the accuracy of the results is increased by the choice of nonlinear activation function  $\sigma$. 

The chosen ODE model has been a \textit{Duffing oscillator} \cite{kovacic2011duffing}
\[
\frac{du}{dx} = \begin{bmatrix}u^{(2)}\\-u^{(1)}-u^{(2)}-0.5\left(u^{(1)}\right)^3\end{bmatrix}.
\]

\begin{figure}[!htb]
	\centering
	\input{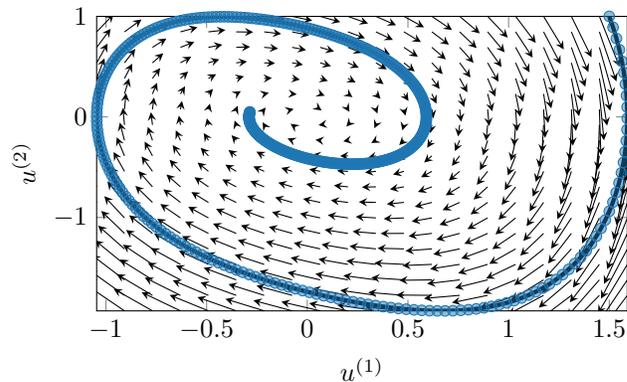}
	\vspace{-5mm}
	\caption{\footnotesize Quiver plot of the vector field of the Duffing ODE described in. The blue points are sampled from a single continuous trajectory and used for the batch training procedure.}
	\label{fig:VecField}
\end{figure}

Given the initial condition $u_0 = [1.5,1]^\top$, a trajectory $u(t)$ was numerically integrated in $x\in [0,8]$ via the {\tt odeint} solver of \textit{Scipy} library and 400 evenly distributed measurements have been collected (i.e., $\delta x = x_{i+1}-x_i = 0.2~\forall i=1,\dots,s$ ). The vector field and the computed trajectory are shown in Fig. \ref{fig:VecField}. 

A three layers neural network has been selected with the two hidden layers having a width  $h_1 = h_2 = 16$. The total number of network parameters is $p = 354$. The design of a network with two hidden layers instead of a single, larger hidden layer or additional, narrower layers is motivated by \cite{lu2017expressive} and \cite{eldan2016power}. While traditionally depth has been regarded as the more important attribute, recent developments have shown that a correct balance of depth and width can be beneficial for neural network performance.

The activation function has been selected as $\sigma(\cdot) \triangleq \frac{1}{\gamma}\ln(1+e^{-\gamma(\cdot}))$ ($\gamma = 10$). This function, referred as \textit{softplus}, is the smooth counterpart of the more popular \textit{ReLu} activation. ReLu offers fast convergence to a minimum due to its linear region but is not differentiable in ${0}$. While in practice this drawback rarely causes problem due to numerical approximation, the choice of softplus was made to not violate the smoothness assumption of $\mathcal{J}^*$.  

\begin{figure}[!t]
	\centering
	\includegraphics{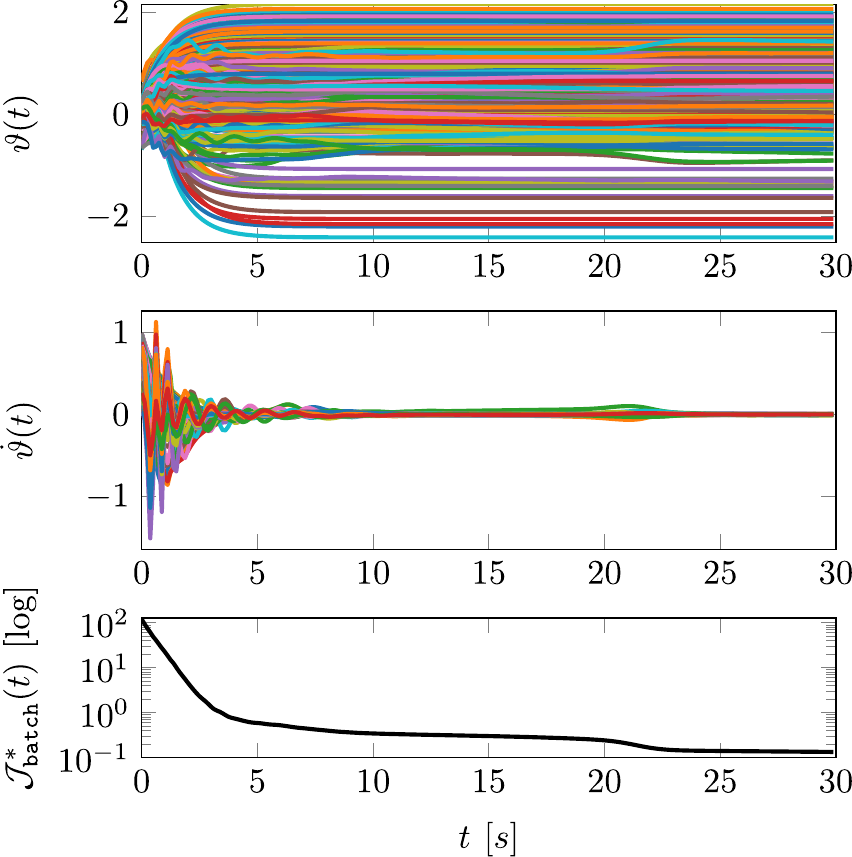}
	\vspace{-9mm}
	\caption{\footnotesize  Results of the \textit{batch} training of the neural network for the vector field reconstruction. [Above] Trajectory of the 354 parameters $\vartheta(t)$ and their velocities $\dot{\vartheta}$. [Below] Decay of the loss function over time.}
	\label{fig:TrajVecField}
\end{figure}

The PH model of the parameters and the objective function have been defined as in Example \ref{exmp:L2pot} with $\alpha = 1,~ \beta = 0$ and $B = 0.5\cdot I_p$. $\vartheta$ and $\dot{\vartheta}$ has been initialized sampling a Gaussian distribution with unitary variance and a uniform distribution over $[0, 1)$ respectively. The training has been performed with \textit{batch} method by numerically integrating the PH model for 100s. The training outcomes of first $30s$ the are shown in Fig. \ref{fig:TrajVecField}. It can be noticed that after $30s$ most of the parameters have converged, thus reaching a minimum of $\mathcal{J}^*_{\tt batch}$. Around the $20s$ point some of the velocities show a ripple, followed by a variation of the corresponding parameters. The loss decay is simultaneously accelerated during this event due to the dissipation term $B\dot{\vartheta}$. This behavior is most likely due to the state passing through a saddle point of the $\mathcal{J}^*_{\tt batch}$.

\begin{figure}[!htb]
	\centering
	\begin{tikzpicture}
\definecolor{color0}{rgb}{0.12156862745098,0.466666666666667,0.705882352941177}
\begin{axis}[
width = \linewidth,
height = 5.5cm,
view={0}{90},
xmin = {-1.04625177},
xmax = { 1.58974179},
ymin = {-1.92847334},
ymax = {1},
colormap = {whiteblack}{color(0cm)  = (white);color(1cm) = (black)},
colorbar horizontal,
colorbar style = {
                 at = {(0,-.25)},
                 ylabel={$\mathcal{E}(u)$}
                 },
xlabel={$u^{(1)}$},
xlabel style = {at = {(0.5,-.075)}},
ylabel={$u^{(2)}$},
axis background/.style={fill=white},
]
]
\addplot3[contour filled={number = 50,labels={false}},mesh/rows=20,mesh/cols=20,mesh/check=false,forget plot
] table {Error3.dat};

\addplot[orange,
quiver={u=\thisrow{u},v=\thisrow{v},scale arrows=0.1},
-stealth] table {VecField.dat};

\addplot[color0,
quiver={u=\thisrow{u},v=\thisrow{v},scale arrows=0.1},
-stealth] table {RecVecField3.dat};

\end{axis}
\end{tikzpicture}
	\vspace{-5mm}
	\caption{\footnotesize Learned vector field (blue arrows) versus true vector field (orange arrows) and absolute reconstruction error.}
	\label{fig:VecFieldErr}
\end{figure}
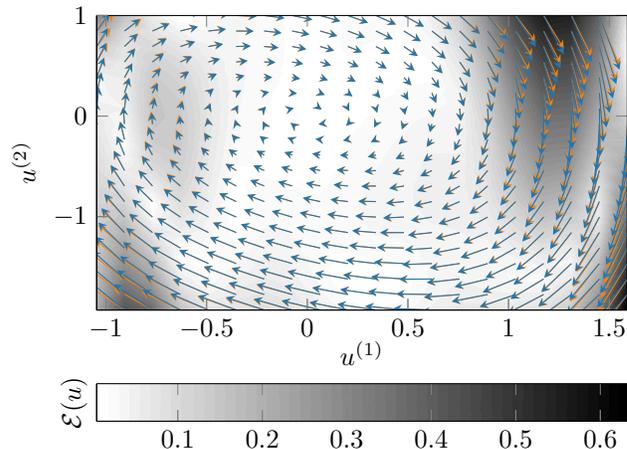

The error $\mathcal{E}(u)$ has then been computed for $u\in\D\triangleq[-1,1.5]\times [-1.9,1]$. Figure \ref{fig:VecFieldErr} shows that the reconstruction error is highest in the state-space regions from which the neural network received no training information. The neural network has been able to infer the shape of the vector field elsewhere, especially in regions with a higher training data density.

%
%
%




\section{CONCLUSIONS}\label{sec:concl}
In this work we provide a new perspective on the process of neural network optimization. Inspired by their modular nature, we design objective function and parameter training dynamics in such a way that the neural network itself behaves as an autonomous Port-Hamiltonian system. A result is the implicit guarantee on convergence to a minimum of the loss function due to PH passivity.
{In the context of training neural networks, escaping from saddle points has been a challenge due to the non--convexity and high--dimensionality of the optimization problem. The proposed framework is promising since it it circumvents the problem of getting stuck at saddle points by guaranteeing convergence to a minimum of the loss function.}
In juxtaposition with the discrete nature of many other popular neural network optimization schemes currently used in state-of-the-art deep learning models, our framework features a continuous evolution of the parameters. Future work will be carried out in order to exploit this property to shed more light on some of the underlying characteristics of neural networks, especially those with a high number of layers, the behavior of which is proving to be quite challenging to model. 
Additionally, this framework enables a treatment of neural networks based on physical systems and PH control which can increase the performance of the learning procedure and the probability of finding the global minimum of the objective function. Here, we performed experiments on classification and vector field approximation and determined that the proposed method scales up to neural networks of non-trivial size.

\bibliographystyle{unsrt}
\bibliography{root_final.bib}

\end{document}